\documentclass[lettersize,journal]{IEEEtran}

\usepackage{amsmath,amsfonts}
\usepackage{algorithmic}
\usepackage{array}
\usepackage[caption=false,font=normalsize,labelfont=sf,textfont=sf]{subfig}
\usepackage{textcomp}
\usepackage{stfloats}
\usepackage{url}
\usepackage{verbatim}
\usepackage{graphicx}
\usepackage[table]{xcolor}
\usepackage{booktabs}
\usepackage{multirow}
\usepackage{tabularx}
\usepackage{amssymb}
\def\BibTeX{{\rm B\kern-.05em{\sc i\kern-.025em b}\kern-.08em
    T\kern-.1667em\lower.7ex\hbox{E}\kern-.125emX}}
\usepackage{balance}
\usepackage{placeins} 
\usepackage{cite} 
\usepackage[colorlinks=false,pdfborder={0 0 0}]{hyperref} 
\newsavebox{\tablemainbox}

\begin{document}

\title{PROSPECT: Unified Streaming Vision-Language Navigation via Semantic--Spatial Fusion and Latent Predictive Representation}

\author{Zehua~Fan$^{1,*}$,
        Wenqi~Lyu$^{3}$,
        Wenxuan~Song$^{5}$,
        Linge~Zhao$^{4}$,
        Yifei~Yang$^{6}$,
        Xi~Wang$^{7}$,
        Junjie~He$^{5}$,
        Lida~Huang$^{2}$,
        Haiyan~Liu$^{8}$,
        Bingchuan~Sun$^{8,\diamond}$,
        Guangjun~Bao$^{8}$,
        Xuanyao~Mao$^{8}$,
        Liang~Xu$^{8}$,
        Yan~Wang$^{2,\diamond}$,
        Feng~Gao$^{1,\diamond}$%
\thanks{$^{1}$Z.~Fan and F.~Gao are with Shanghai Jiao Tong University, Shanghai, China. $^{2}$L.~Huang and Y.~Wang are with Tsinghua University, Beijing, China. $^{3}$W.~Lyu is with the University of Adelaide, Adelaide, Australia. $^{4}$L.~Zhao is with Wuhan University, Wuhan, China. $^{5}$W.~Song and J.~He are with the Hong Kong University of Science and Technology (Guangzhou), Guangzhou, China. $^{6}$Y.~Yang is with Beijing Jiaotong University, Beijing, China. $^{7}$X.~Wang is with AIR Wuxi Innovation Center, Tsinghua University, Wuxi, China. $^{8}$H.~Liu, B.~Sun, G.~Bao, X.~Mao, and L.~Xu are with Lenovo, Beijing, China. $^{\diamond}$\,Corresponding authors: Yan~Wang, Bingchuan~Sun and Feng~Gao.}%
}


\maketitle

\begin{abstract}
Multimodal large language models (MLLMs) have advanced zero-shot end-to-end Vision-Language Navigation (VLN), yet robust navigation requires not only semantic understanding but also predictive modeling of environment dynamics and spatial structure. We propose \emph{PROSPECT}, a unified streaming navigation agent that couples a streaming Vision-Language-Action (VLA) policy with latent predictive representation learning. PROSPECT uses CUT3R as a streaming 3D foundation spatial encoder to produce long-context, absolute-scale spatial features, and fuses them with SigLIP semantic features via cross-attention. During training, we introduce learnable \emph{stream query tokens} that query the streaming context and predict next-step 2D and 3D latent features (rather than pixels or explicit modalities), supervised in the latent spaces of frozen SigLIP and CUT3R teachers. The predictive branch shapes internal representations without inference overhead. Experiments on VLN-CE benchmarks and real-robot deployment demonstrate state-of-the-art performance and improved long-horizon robustness under diverse lighting. \emph{We will release code for the community soon.}
\end{abstract}

\begin{IEEEkeywords}
Vision-language navigation, embodied AI, world models.
\end{IEEEkeywords}

\section{Introduction}

\IEEEPARstart{V}{ision-Language} Navigation (VLN) is a key step toward general-purpose embodied agents. Recent multimodal large language models (MLLMs) enable strong zero-shot VLN by mapping egocentric observations to actions in a Vision-Language-Action (VLA) paradigm \cite{CIT:Navid,zhang2024uni,cheng2024navila,wei2025streamvln}.

Nevertheless, strong navigation performance depends not only on understanding the world but also on predicting and generating future outcomes. World models address this by predicting future states from past context \cite{assran2023self,assran2025v,wang2024emu3,wan2025wan,bar2025navigation}. In VLA, unified frameworks that learn both action generation and predictive representations \cite{worldvla2025,rynnvla0022025,dreamvla2025} can be more compact and synergistic than combining an MLLM with a separate video generator \cite{pan2025transfer,mantis2025,motus2025}. In navigation, existing predictive approaches either rely on low-dimensional state-space models with limited expressivity \cite{yao2025navmorph}, or supervise in explicit pixel/depth spaces (concurrent work \cite{liu2025navforesee}), which may overfit to task-irrelevant details such as textures and illumination, degrading out-of-domain robustness. Many prior models condition on short history \cite{dreamvla2025,mantis2025,motus2025}, underutilizing long streaming context.

Streaming VLN benefits from long-range context. StreamVLN \cite{wei2025streamvln} introduces a fast--slow context mechanism, but remains a VLA-only method without an explicit predictive component under streaming RGB.

The vision encoder also shapes downstream performance. Many VLN methods rely on 2D semantic encoders (e.g., SigLIP \cite{tschannen2025siglip}) and thus lack spatial intelligence. Recent 3D foundation models (VGGT \cite{wang2025vggt} and successors \cite{zhuo2025streaming,yuan2026infinitevggt}, and CUT3R \cite{wang2025continuous}) extract spatial features from RGB and are often fused with 2D features \cite{spatialmllm2025,vlm3r2025,spatialforcing2025}. In VLN, a concurrent line of work
explores VGGT as a spatial encoder \cite{janusvln2025}. VGGT-based encoders can be memory-heavy for long episodes and require ad-hoc history truncation to avoid out-of-memory (OOM) at inference. They also provide relative-scale representations, complicating the maintenance of consistency under large viewpoint changes. CUT3R is inherently streaming and yields absolute-scale spatial features, making it suited for long-context streaming navigation.

\begin{figure*}[!t]
\centering
\includegraphics[width=\textwidth]{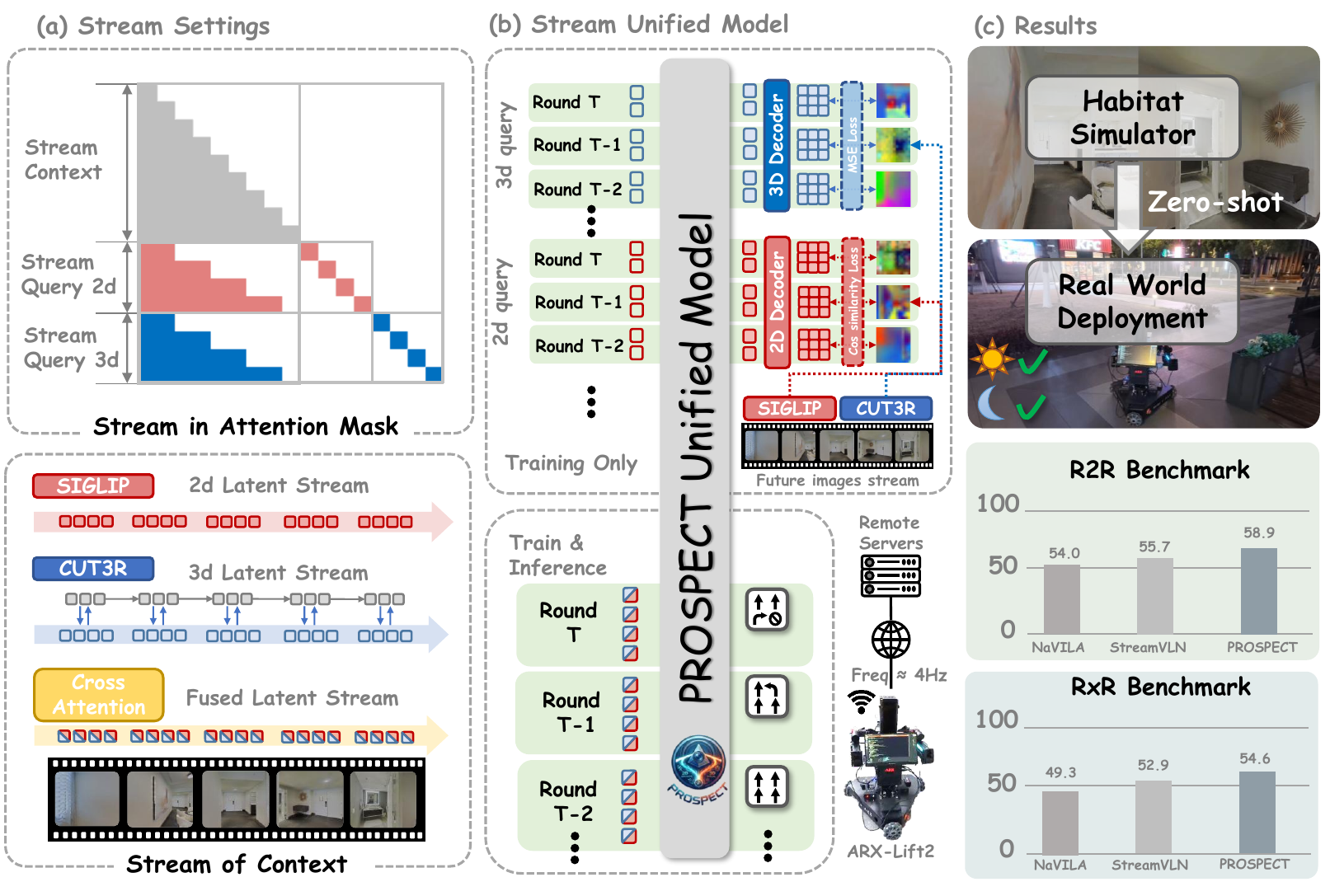}
\caption{Overview of PROSPECT. (a) \textbf{Streaming setup:} A streaming attention mask enforces temporal causality and isolates 2D/3D query tokens to prevent cross-modal leakage. SigLIP and CUT3R provide 2D semantic and absolute-scale 3D spatial feature streams, fused by cross-attention for the policy. (b) \textbf{Unified model:} In training, stream query tokens predict next-step 2D/3D latent features under frozen SigLIP/CUT3R supervision (no inference cost). At inference, only the VLA policy runs at $\sim$4\,Hz. (c) \textbf{Results:} First-tier VLN-CE performance and zero-shot Habitat navigation; larger gains on the long-horizon RxR benchmark than on R2R, indicating stronger robustness for complex instruction following. Real-robot deployment is robust under diverse lighting.}
\label{fig:overview}
\end{figure*}

To address these limitations, we propose \textbf{PROSPECT} (\textbf{P}redictive \textbf{R}epresentations \textbf{O}f \textbf{SP}atial-s\textbf{E}mantic \textbf{C}ontex\textbf{T}s), a unified architecture that combines streaming VLA with latent-space predictive representation learning, featuring long contextual
semantics, spatial understanding, and predictive capabilities. PROSPECT takes streaming video and uses the streaming 3D foundation model (CUT3R) to continuously encode it into spatial features with absolute scale, which enhance 2D semantic features. Inspired by JEPA \cite{assran2023self,assran2025v}, we predict future \emph{latent} 2D/3D features rather than pixels/depth. We introduce \emph{stream query tokens} to query the long streaming context and decode next-step 2D semantic and 3D spatial latent features during training. At inference, the predictive branch is removed; it has shaped the VLA representations to internalize dynamics without adding latency. Fig.~\ref{fig:overview} overviews the method.

Our contributions are:
\begin{itemize}
\item A unified streaming VLN framework that integrates streaming VLA with latent predictive representation learning, achieving first-tier VLN-CE performance.
\item CUT3R-based streaming 3D perception with absolute-scale spatial features for efficient long-context navigation.
\item Stream query tokens with a streaming-causal attention mask that enables latent prediction while disentangling 2D/3D objectives.
\item Real-robot deployment demonstrating high-frequency control and robustness across indoor and outdoor scenes under diverse lighting conditions.
\end{itemize}

\section{Related Works}

\subsection{Embodied Navigation}
Embodied navigation spans Vision Navigation (VN) and VLN. Earlier VLN systems established benchmarks and reliable continuous-environment protocols \cite{krantz2021waypoint, hong2022bridging, krantz2022sim, wang2023gridmm, an2024etpnav, long2024instructnav, chen2021topological, wang2024sim}. With MLLMs and VLA models, approaches represented by NaVid advance end-to-end navigation from egocentric RGB without odometry or pre-built maps \cite{CIT:Navid,cheng2024navila,zhang2024uni,wei2025streamvln,yu2025correctnav}. Nevertheless, they primarily emphasize action generation and language grounding, leaving spatial understanding and future prediction capability underexplored in a unified streaming setting. We propose a unified streaming VLN paradigm that couples spatial understanding with latent predictive representation learning via stream query tokens, yielding an end-to-end agent that handles long-context streams, maintains spatial grounding, and remains prediction-shaped.

\subsection{World Models and Predictive Representations}
World models are often framed as future frame generators \cite{liu2024sora,wan2025wan}, and in embodied settings extend to explicit modalities (BEV/occupancy/depth/segmentation) \cite{zhou2025hermes,dang2025sparseworld,dreamvla2025,liu2025navforesee}. Supervision in explicit spaces can overweight task-irrelevant appearance factors. In contrast, motivated by JEPA \cite{assran2023self,assran2025v}, we supervise prediction directly in compact latent spaces of 2D semantics and 3D spatial features, encouraging dynamics-aware representations without modeling pixel noise.

In VLN, few works combine future frame or latent feature prediction with navigation \cite{yao2025navmorph,liu2025navforesee}; they often omit 3D fusion or depend on simulator-provided poses and ground-truth states, thereby limiting mapless and odometry-free deployment. PROSPECT unifies streaming updates, 2D--3D fusion, and latent prediction in an end-to-end framework.

\subsection{Spatial Intelligence}
Spatial intelligence concerns representing and reasoning about 3D structure \cite{yang2025thinking,yang2025cambrian}. Representations include depth/point clouds, 3D Gaussians \cite{kerbl20233d}, BEV/occupancy \cite{zhou2025hermes,dang2025sparseworld}, and 3D foundation features \cite{wang2025vggt,wang2025continuous,zhuo2025streaming,yuan2026infinitevggt,chen2025ttt3r}. 3D foundation features are compact and effective in VLA settings \cite{vlm3r2025,spatialmllm2025}. We adopt CUT3R for its inherent streaming capability and absolute-scale spatial representations, which enable stable long-context VLN.

\section{Method}

\subsection{Problem Formulation: Streaming VLA for VLN}
Streaming VLN can be formulated as a streaming VLA problem.
Given a language instruction $I$, at each time step $t$ the agent receives an observation $o_t$ and produces an action $a_t$, interacting with the environment
in an alternating perception--action stream~\cite{wei2025streamvln}. Let $\mathcal{W}_t = \{o_{t-N+1},\,a_{t-N+1},\,\ldots,\,o_{t-1},\,a_{t-1}\}$
denote the $N{-}1$ preceding observation--action pairs in the $N$ step sliding window. The streaming context is:
\begin{equation}
  \mathrm{Stream}_{0:t} \;:=\; \bigl\{\,\mathrm{KV}(\mathcal{W}_t),\; o_t,\; M\,\bigr\},
  \label{eq:stream}
\end{equation}
where $\mathrm{KV}(\cdot)$ caches the key--value states of the short-term sliding window,
and $M$ is long-term memory tokens summarizing uniformly sampled historical keyframes.

In mapless, odometry-free VLN, $o_t \in \mathbb{R}^{3 \times H \times W}$ is a single-view RGB image. We use standard atomic actions \cite{CIT:Navid,wei2025streamvln}: the model outputs $n_a$ actions per step, $a_t=(a_t^{(1)},\ldots,a_t^{(n_a)})$ with $n_a=4$, where
\begin{equation}
\label{eq:action_space}
a_t^{(i)} \in \mathcal{A} := \{\uparrow,\ \leftarrow,\ \rightarrow,\ \texttt{STOP}\}.
\end{equation}
Here $\uparrow$ is moving forward 25\,cm and $\leftarrow/\rightarrow$  are turning 15$^\circ$ left/right. The streaming policy is
\begin{equation}
\label{eq:vla}
a_t = \text{VLA}(I,\ \text{Stream}_{0:t}).
\end{equation}

\subsection{Unified Streaming Navigation with Latent Prediction}
Temporal correlations among semantics, spatial layout, physical dynamics, and task progress are rich yet implicit. Next-frame prediction is naturally compatible with streaming and can benefit from scaling, while improving physical understanding. We thus propose a unified streaming model that produces both navigation actions and future latent features:
\begin{equation}
\label{eq:um}
a_t,\ \mathbf{F}^{2\text{D}}_{t+1},\ \mathbf{F}^{3\text{D}}_{t+1} = \text{UM}(I,\ \text{Stream}_{0:t}),
\end{equation}
where $\text{UM}(\cdot)$ denotes PROSPECT, and $\mathbf{F}^{2\text{D}}_{t+1}$/$\mathbf{F}^{3\text{D}}_{t+1}$ are predicted 2D semantic / 3D spatial latent features.

\begin{figure*}[!t]
\centering
\includegraphics[width=\textwidth]{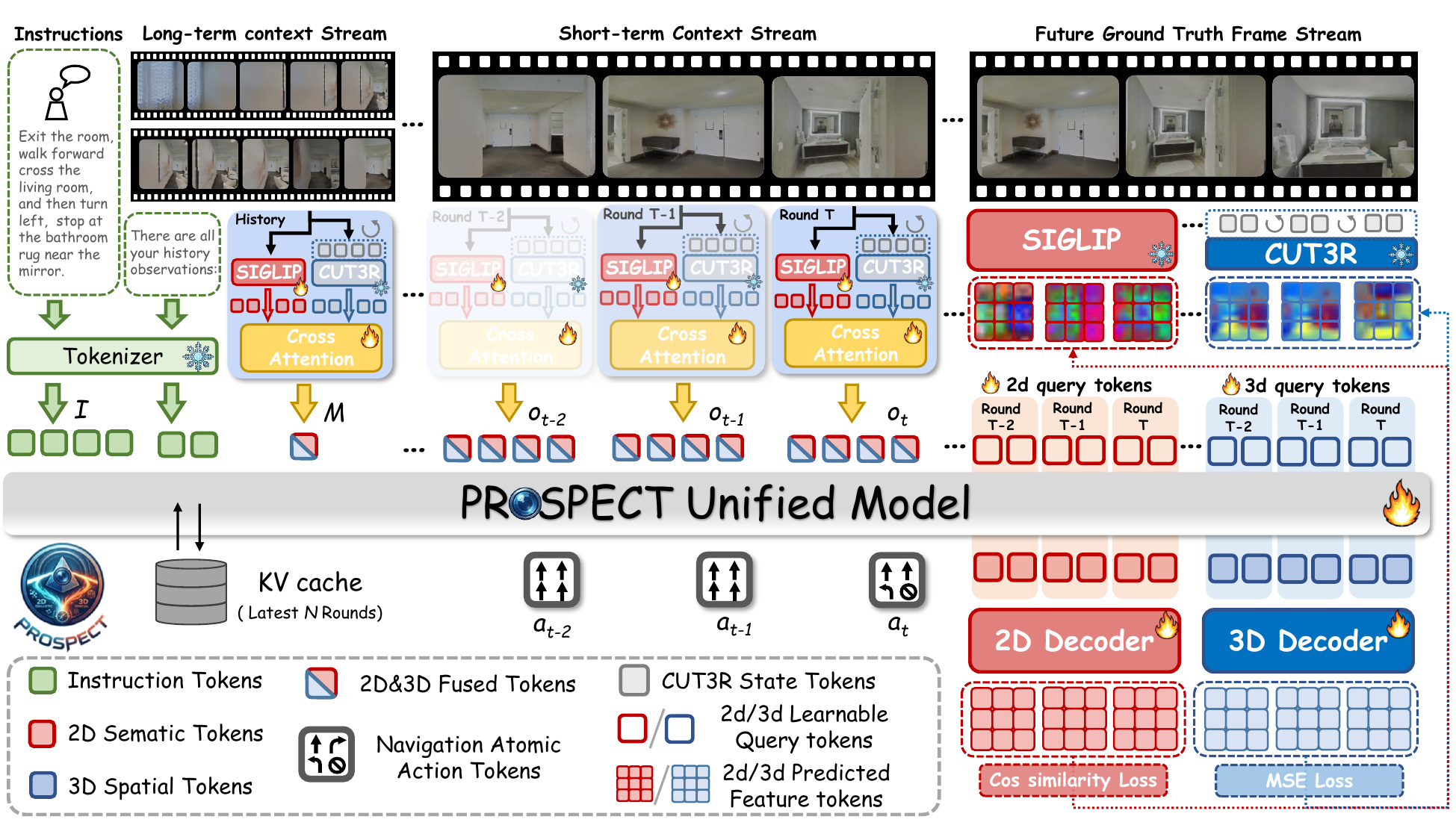}
\caption{Architecture of PROSPECT. Instruction and observations (historical keyframes and current frame) share one pipeline: frozen SigLIP and CUT3R with cross-attention fusion; keyframes are condensed into long-term memory $M$. The model uses a KV cache for context and autoregressively outputs navigation actions. Training only: 2D/3D query tokens reverse-query the stream; lightweight decoders predict next-step latents under cosine (2D) and MSE (3D) with frozen teachers. Predictive branch removed at inference.}
\label{fig:architecture}
\end{figure*}

Fig.~\ref{fig:architecture} shows the architecture. The VLA branch encodes fused 2D--3D features and autoregressively outputs actions. During training, the predictive branch appends time-ordered query tokens that attend to the streaming context; lightweight decoders then predict next-step latent features. Understanding and generation are unified in one streaming framework.

\subsection{Perception and Representation: 2D--3D Fusion}
SigLIP encodes each observation into 2D semantic features:
\begin{equation}
\label{eq:siglip}
\mathbf{F}^{2\text{D}}_t = \text{SigLIP}(o_t).
\end{equation}

For spatial features, we use CUT3R as a streaming 3D encoder. CUT3R first encodes the frame by a ViT encoder,
\begin{equation}
\label{eq:cut3r_pre}
\mathbf{F}^{3\text{D,pre}}_t = \text{Encoder}(o_t).
\end{equation}
With a previous state token $\mathbf{s}_{t-1}$, a learnable pose token $\mathbf{p}_t$, and current features $\mathbf{F}^{3\text{D,pre}}_t$, the decoder rolls out spatial features and updates the state:
\begin{equation}
\label{eq:cut3r_rollout}
[\mathbf{p}'_t,\ \mathbf{F}^{3\text{D}}_t],\ \mathbf{s}_t
= \text{Decoders}([\mathbf{p}_t,\ \mathbf{F}^{3\text{D,pre}}_t],\ \mathbf{s}_{t-1}).
\end{equation}

We fuse $\mathbf{F}^{2\text{D}}_t$ and $\mathbf{F}^{3\text{D}}_t$ via cross-attention:
\begin{equation}
\label{eq:cross_attn}
\mathbf{F}^{\text{fuse}}_t
=
\text{softmax}\!\left(
\frac{(\mathbf{F}^{2\text{D}}_t \mathbf{W}_Q)(\mathbf{F}^{3\text{D}}_t \mathbf{W}_K)^\top}{\sqrt{d_k}}
\right)
(\mathbf{F}^{3\text{D}}_t \mathbf{W}_V),
\end{equation}
where $d_k$ is the key dimension and $\mathbf{W}_Q,\mathbf{W}_K,\mathbf{W}_V$ are learnable projections. Each $\mathbf{F}^{\text{fuse}}_t$ is then mapped through an MLP into the LLM embedding space and fed to the LLM along with instruction tokens. The historically sampled keyframes that form long-term memory $M$ are encoded with the same 2D--3D fusion pipeline; each keyframe's fused representation is condensed into a single token before being fed to the LLM.

\subsection{Latent Prediction via Stream Query Tokens}
While fused 2D--3D features provide a \emph{forward} aggregation of streaming information into the LLM, we introduce stream query tokens to \emph{reverse-query} the streaming context and predict future latent features. At step $t$, we append learnable tokens $\langle q^{2\text{D}}_t\rangle$ and $\langle q^{3\text{D}}_t\rangle$ to the LLM input, yielding compact embeddings of the future time step $t+1$:
\begin{align}
\label{eq:query_embed_2d}
\mathbf{e}^{2\text{D}}_{t+1} &= \text{LLM}(I,\ \text{Stream}_{0:t}\ |\ \langle q^{2\text{D}}_t\rangle),\\
\label{eq:query_embed_3d}
\mathbf{e}^{3\text{D}}_{t+1} &= \text{LLM}(I,\ \text{Stream}_{0:t}\ |\ \langle q^{3\text{D}}_t\rangle).
\end{align}
Two lightweight Transformer decoders reconstruct token-level latent features from these embeddings:
\begin{align}
\label{eq:decode_2d}
\widehat{\mathbf{F}}^{2\text{D}}_{t+1} &= \text{Decoder}_{2\text{D}}(\mathbf{e}^{2\text{D}}_{t+1}\ |\ \langle m_t^{2\text{D}}\rangle),\\
\label{eq:decode_3d}
\widehat{\mathbf{F}}^{3\text{D}}_{t+1} &= \text{Decoder}_{3\text{D}}(\mathbf{e}^{3\text{D}}_{t+1}\ |\ \langle m_t^{3\text{D}}\rangle),
\end{align}
where $\langle m_t^{2\text{D}}\rangle$ and $\langle m_t^{3\text{D}}\rangle$ are learnable masked tokens repeated to match the target token length. Each decoder has 2 layers and predicts a full-length latent sequence aligned with the target image-token sequence.

Targets $\mathbf{F}^{2\text{D}}_{t+1}$ and $\mathbf{F}^{3\text{D}}_{t+1}$ are computed from the next-step observation using frozen SigLIP and CUT3R teachers (no gradient). We supervise 2D with cosine distance and 3D with MSE:
\begin{align}
\label{eq:l2d}
\mathcal{L}_{2\text{D}} &= 1 - \cos\!\left(\widehat{\mathbf{F}}^{2\text{D}}_{t+1},\ \mathbf{F}^{2\text{D}}_{t+1}\right),\\
\label{eq:l3d}
\mathcal{L}_{3\text{D}} &= \text{MSE}\!\left(\widehat{\mathbf{F}}^{3\text{D}}_{t+1},\ \mathbf{F}^{3\text{D}}_{t+1}\right).
\end{align}
SigLIP is trained with a pairwise sigmoid loss on \emph{$\ell_2$-normalized} embeddings. In our experiments, cosine loss aligns well with this normalized geometry, whereas applying MSE to 2D features penalizes norm differences and leads to unstable training; in contrast, MSE was stable for CUT3R features.

The overall objective is
\begin{equation}
\label{eq:loss_all}
\mathcal{L}_{\text{all}}
=
\mathcal{L}_{\text{nav}}
+
\gamma\left(\alpha\,\mathcal{L}_{2\text{D}} + \beta\,\mathcal{L}_{3\text{D}}\right),
\end{equation}
where $\mathcal{L}_{\text{nav}}$ is action cross-entropy \cite{wei2025streamvln} and $(\gamma,\alpha,\beta)$ are set in Sec.~\ref{sec:training}.

\subsection{Streaming Attention Mask}
A standard causal mask is insufficient because we introduce per-step predictive queries. We interpret the short-term navigation context as an $N$-turn dialogue: at each turn $i$, the model consumes context $\text{ctxt}_i$ (prompt and observation tokens) and produces a response $\text{act}_i$ (actions). The initial turn includes the instruction and long-term memory $M$. During training, we augment each turn by appending $\langle q^{2\text{D}}_i\rangle$ and $\langle q^{3\text{D}}_i\rangle$ to the end of the input sequence.

\begin{figure}[!t]
\centering
\includegraphics[width=\linewidth]{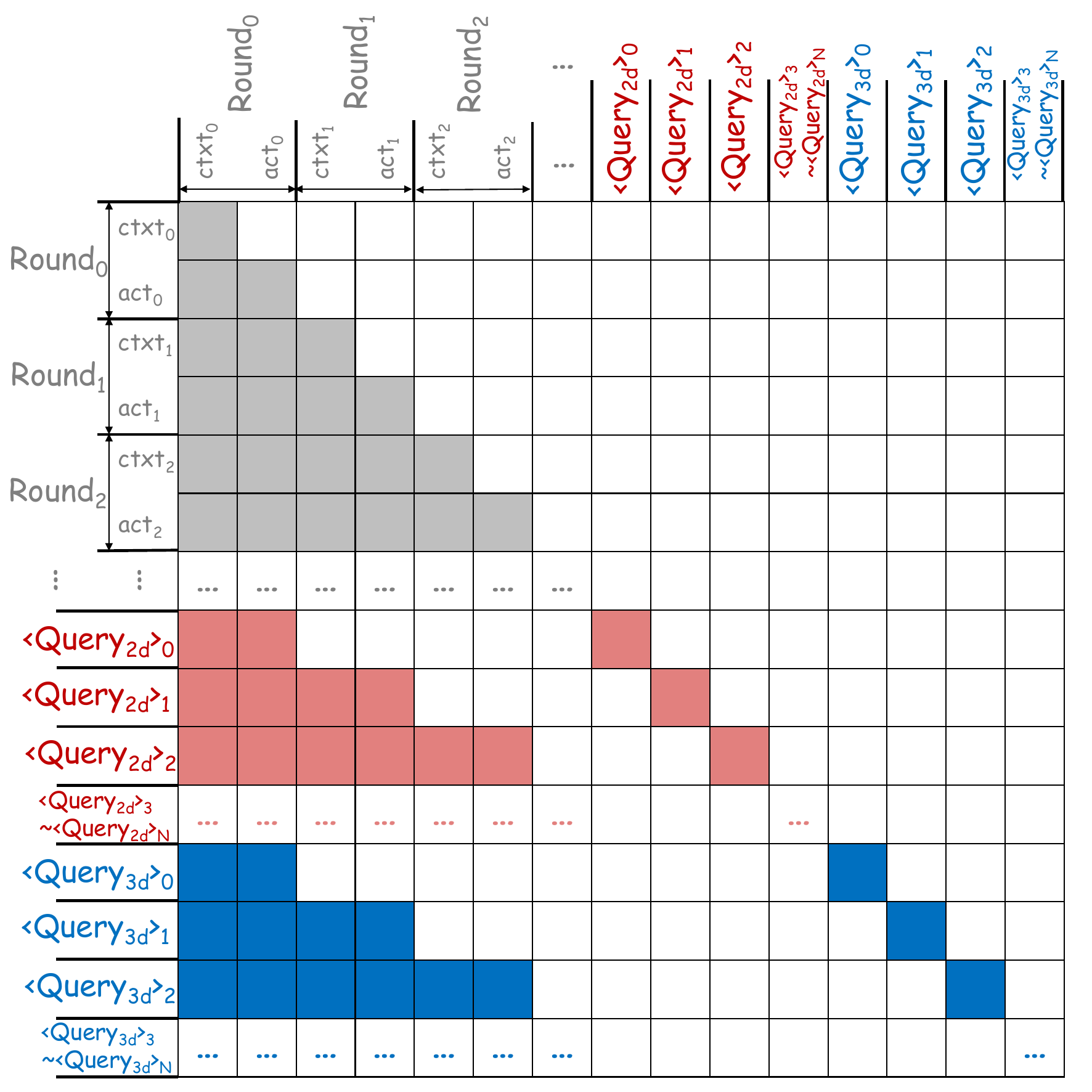}
\caption{Streaming attention mask used by PROSPECT. \textbf{Upper (gray):} Causal mask for navigation context (ctxt) and actions (act): each $\text{act}_i$ may attend only to $\text{ctxt}_{0:i}$ and $\text{act}_{0:i-1}$, ensuring no future leakage. \textbf{Middle (red):} Each 2D query token $\langle\text{Query2d}_i\rangle$ attends only to its own round and prior rounds' ctxt/act; it cannot attend to any other Query2d, any Query3d, or future rounds---enforcing both turn isolation and modality disentanglement. \textbf{Lower (blue):} Same for 3D query tokens $\langle\text{Query3d}_i\rangle$.}
\label{fig:attnmask}
\end{figure}

We enforce three constraints for correct causality, learning efficiency, and train--test alignment in the unified streaming model. First, for causality, each query token attends only to its own turn and all previous turns, never to future turns. Second, to avoid leakage and reduce error accumulation, query tokens from different turns are isolated (no mutual attention), so each query extracts information only from the shared streaming context rather than from other queries. Third, to disentangle modality-specific supervision, 2D and 3D queries are mutually masked and cannot attend to each other, reducing cross-task interference and preventing degenerate information mixing. During evaluation, we remove the query-token prediction branch; the remaining token sequence preserves the same relative ordering and attention structure as in training, ensuring inference efficiency and effective use of the prediction-shaped representations learned during training. Fig.~\ref{fig:attnmask} illustrates the mask.

\section{Experimental Setup}
\label{sec:training}

\subsection{Datasets and Training Settings}
We adopt StreamVLN \cite{wei2025streamvln} as baseline and use LLaVA-NeXT-Video-7B with a Qwen1.5-7B LLM \cite{ahmed2025qwen}. Following \cite{wei2025streamvln}, we set short-term window $N=8$ and sample 8 long-term keyframes for $M$. Training uses 8$\times$ A800 GPUs in two stages.

\textbf{Stage 1 (SFT).} One-epoch supervised fine-tuning on VLN-CE data in Matterport3D (MP3D): R2R \cite{anderson2018vision}, RxR \cite{ku2020room}, and R2R-EnvDrop \cite{tan2019learning} (total $\sim$479K; R2R/RxR/EnvDrop contribute $\sim$5\%/$\sim$14\%/$\sim$80\%). One epoch costs 560 A800 GPU-hours.

\textbf{Stage 2 (Augmented SFT).} We retain the Stage~1 R2R/RxR trajectories to mitigate forgetting, and add \(\sim\)260K DAgger samples \cite{ross2011reduction}, where expert relabeling provides recovery actions for off-policy deviations, as well as \(\sim\)314K ScaleVLN samples \cite{ramakrishnan2021habitat,wang2023scaling}, a large-scale VLN dataset in Habitat Matterport3D (HM3D). To encourage multimodal and spatial reasoning, we mix in VQA data targeting video-based spatial and geometric understanding, including LLaVA-Video-178K \cite{zhang2410video} and ScanQA \cite{azuma2022scanqa}. Stage~2 mixture contains \(\sim\)938K samples (71\% VLN, 29\% VQA), and one epoch costs \(\sim\)1900 A800 GPU-hours.

We set the learning rate to \(5\times10^{-6}\) for SigLIP and use a peak \(2\times10^{-5}\) for all other trainable modules; CUT3R is frozen. Warm-up ratios are 7.5\% (Stage~1) and 3\% (Stage~2). We set \(\gamma=0.01\), \(\alpha=0.25\), and \(\beta=0.75\) to balance the loss scales, preventing any single term from dominating due to magnitude alone. We use 196 masked tokens and 9 query tokens per modality, trading off representation capacity, the LLM input token budget, and compute cost.

\subsection{Benchmarks and Metrics}
We evaluate on VLN-CE in Habitat \cite{krantz2020beyond}, reporting Success Rate (SR), Success weighted by Path Length (SPL), Navigation Error (NE), and Oracle Success Rate (OSR) on the \texttt{val-unseen} split of R2R and RxR \cite{anderson2018vision,ku2020room}.

\subsection{Real-Robot Deployment}
We deploy on an ARX-Lift2 robot using egocentric RGB from a head-mounted RealSense 405. As in most prior VLN robot deployments \cite{CIT:Navid,wei2025streamvln}, we run remote inference over Wi-Fi/LAN indoors (dual RTX-4090 server; $\sim$0.25\,s/step, $\sim$4\,Hz) and over public network outdoors (dual A800 server; $\sim$0.27\,s/step, $\sim$4\,Hz) with authenticated access. We also test onboard inference on a single RTX 4070 with reduced precision; success is lower but remains feasible for less demanding scenarios.

\section{Experiments}
\begin{table*}[!t]
\centering
\caption{Comparison with state-of-the-art methods on VLN-CE R2R and RxR Val-Unseen split.}
\label{tab:main}
\footnotesize
\setlength{\tabcolsep}{5pt}
\sbox{\tablemainbox}{%
\begin{tabularx}{\textwidth}{@{}p{4.1cm}@{\hspace{6pt}}*{4}{>{\centering\arraybackslash}X}*{8}{>{\raggedleft\arraybackslash}X}@{}}
\toprule
\multirow{2}{*}{Method} & \multicolumn{4}{c}{Obs. Enc.} & \multicolumn{4}{c}{R2R Val-Unseen} & \multicolumn{4}{c}{RxR Val-Unseen} \\
\cmidrule(lr){2-5} \cmidrule(lr){6-9} \cmidrule(lr){10-13}
& Pano. & Odo. & D. & S.RGB & NE$\downarrow$ & OSR$\uparrow$ & SR$\uparrow$ & SPL$\uparrow$ & NE$\downarrow$ & SR$\uparrow$ & SPL$\uparrow$ & nDTW$\uparrow$ \\
\midrule
HPN+DN \textcolor{gray}{[ICCV21]} \cite{krantz2021waypoint} & $\checkmark$ & $\checkmark$ & $\checkmark$ & -- & 6.31 & 40.0 & 36.0 & 34.0 & -- & -- & -- & -- \\
CMA \textcolor{gray}{[CVPR22]} \cite{hong2022bridging} & $\checkmark$ & $\checkmark$ & $\checkmark$ & -- & 6.20 & 52.0 & 41.0 & 36.0 & 8.76 & 26.5 & 22.1 & 47.0 \\
VLN$\circlearrowleft$BERT \textcolor{gray}{[CVPR22]} \cite{hong2022bridging} & $\checkmark$ & $\checkmark$ & $\checkmark$ & -- & 5.74 & 53.0 & 44.0 & 39.0 & 8.98 & 27.0 & 22.6 & 46.7 \\
Sim2Sim \textcolor{gray}{[ECCV22]} \cite{krantz2022sim} & $\checkmark$ & $\checkmark$ & $\checkmark$ & -- & 6.07 & 52.0 & 43.0 & 36.0 & -- & -- & -- & -- \\
GridMM \textcolor{gray}{[ICCV23]} \cite{wang2023gridmm} & $\checkmark$ & $\checkmark$ & $\checkmark$ & -- & 5.11 & 61.0 & 49.0 & 41.0 & -- & -- & -- & -- \\
\midrule
InstructNav \textcolor{gray}{[arXiv24]} \cite{long2024instructnav} & -- & -- & -- & -- & 6.89 & -- & 31.0 & 24.0 & -- & -- & -- & -- \\
AG-CMTP \textcolor{gray}{[CVPR21]} \cite{chen2021topological} & $\checkmark$ & $\checkmark$ & $\checkmark$ & -- & 7.90 & 39.2 & 23.1 & 19.1 & -- & -- & -- & -- \\
R2R-CMTP \textcolor{gray}{[CVPR21]} \cite{chen2021topological} & $\checkmark$ & $\checkmark$ & $\checkmark$ & -- & 7.90 & 38.0 & 26.4 & 22.7 & -- & -- & -- & -- \\
LAW \textcolor{gray}{[EMNLP21]} \cite{raychaudhuri2021language} & -- & $\checkmark$ & -- & $\checkmark$ & 6.83 & 44.0 & 35.0 & 31.0 & 10.90 & 8.0 & 8.0 & 38.0 \\
CM2 \textcolor{gray}{[CVPR22]} \cite{Georgakis_2022_CVPR} & -- & $\checkmark$ & $\checkmark$ & $\checkmark$ & 7.02 & 41.5 & 34.3 & 27.6 & -- & -- & -- & -- \\
WS-MGMap \textcolor{gray}{[NeurIPS22]} \cite{chen2022weakly} & -- & $\checkmark$ & $\checkmark$ & $\checkmark$ & 6.28 & 47.6 & 38.9 & 34.3 & -- & -- & -- & -- \\
ETPNav+FF \textcolor{gray}{[arXiv24]} \cite{wang2024sim} & -- & $\checkmark$ & $\checkmark$ & $\checkmark$ & 5.95 & 55.8 & 44.9 & 30.4 & 8.79 & 25.5 & 18.1 & -- \\
Seq2Seq \textcolor{gray}{[ECCV20]} \cite{krantz2020beyond} & -- & -- & $\checkmark$ & $\checkmark$ & 7.77 & 37.0 & 25.0 & 22.0 & 12.10 & 13.9 & 11.9 & 30.8 \\
CMA \textcolor{gray}{[ECCV20]} \cite{krantz2020beyond} & -- & -- & $\checkmark$ & $\checkmark$ & 7.37 & 40.0 & 32.0 & 30.0 & -- & -- & -- & -- \\
NavMorph \textcolor{gray}{[ICCV25]} \cite{yao2025navmorph} & -- & -- & $\checkmark$ & $\checkmark$ & 5.75 & 56.9 & 47.9 & 33.2 & 8.85 & 30.8 & 22.8 & 44.2 \\
\midrule
NaVid \textcolor{gray}{[RSS24]} \cite{CIT:Navid} & -- & -- & -- & $\checkmark$ & 5.47 & 49.1 & 37.4 & 35.9 & -- & -- & -- & -- \\
Uni-Navid \textcolor{gray}{[RSS25]} \cite{zhang2024uni} & -- & -- & -- & $\checkmark$ & 5.58 & 53.3 & 47.0 & 42.7 & 6.24 & 48.7 & 40.9 & -- \\
NaVILA \textcolor{gray}{[RSS25]} \cite{cheng2024navila} & -- & -- & -- & $\checkmark$ & 5.37 & 57.6 & 49.7 & 45.5 & -- & -- & -- & -- \\
StreamVLN$^*$ \textcolor{gray}{[arXiv25]} \cite{wei2025streamvln} & -- & -- & -- & $\checkmark$ & 5.47 & 57.8 & 50.8 & 45.7 & 6.72$^\ddagger$ & 48.6$^\ddagger$ & 42.5$^\ddagger$ & 60.2$^\ddagger$ \\
\rowcolor{gray!20}\textbf{PROSPECT (Ours)}$^*$ & -- & -- & -- & $\checkmark$ & 5.31 & 60.3 & 52.0 & 46.2 & 5.93\phantom{$^\ddagger$} & 52.7\phantom{$^\ddagger$} & 42.8\phantom{$^\ddagger$} & 60.6\phantom{$^\ddagger$} \\
\midrule
NaVILA$^\dagger$ \textcolor{gray}{[RSS25]} \cite{cheng2024navila} & -- & -- & -- & $\checkmark$ & 5.22 & 62.5 & 54.0 & 49.0 & 6.77\phantom{$^\ddagger$} & 49.3\phantom{$^\ddagger$} & 44.0\phantom{$^\ddagger$} & 58.8\phantom{$^\ddagger$} \\
StreamVLN$^\dagger$ \textcolor{gray}{[arXiv25]} \cite{wei2025streamvln} & -- & -- & -- & $\checkmark$ & 5.10 & 64.0 & 55.7 & 50.9 & 6.22\phantom{$^\ddagger$} & 52.9\phantom{$^\ddagger$} & 46.0\phantom{$^\ddagger$} & 61.9\phantom{$^\ddagger$} \\
\rowcolor{gray!20}\textbf{PROSPECT (Ours)}$^\dagger$ & -- & -- & -- & $\checkmark$ & \textbf{4.92} & \textbf{65.2} & \textbf{58.9} & \textbf{54.0} & \textbf{5.70}\phantom{$^\ddagger$} & \textbf{54.6}\phantom{$^\ddagger$} & \textbf{46.2}\phantom{$^\ddagger$} & \textbf{62.1}\phantom{$^\ddagger$} \\
\bottomrule
\end{tabularx}}%
\usebox{\tablemainbox}
\par\vspace*{0.5em}
\centerline{\begin{minipage}{\wd\tablemainbox}
\footnotesize
\textbf{Note.} Obs.\ Enc.: Pano.=panoramic, Odo.=odometry, D.=depth, S.RGB=single-view RGB.\\
$^*$: MP3D + VideoQA only (StreamVLN \cite{wei2025streamvln} Table~3 row~2).\\
$^\dagger$: Non-MP3D and Extra data: Ours and StreamVLN add ScaleVLN and MMC4 (\cite{wei2025streamvln} Table~3 row~4); NaVILA adds human-following data.\\
$^\ddagger$: RxR under $^*$ not reported by StreamVLN; we quote their numbers from a recipe that adds MMC4 on top of $^*$ for reference (\cite{wei2025streamvln} Table~1 row~20).
\end{minipage}}
\end{table*}

\begin{table}[!t]
\centering
\caption{Module ablation on R2R \texttt{val-unseen}.}
\label{tab:ablation_modules}
\begin{tabular}{l|c c c c}
\hline
Setting & NE$\downarrow$ & OSR$\uparrow$ & SR$\uparrow$ & SPL$\uparrow$ \\
\hline
Baseline (SigLIP only) & 6.05 & 53.8 & 45.5 & 41.6 \\
Ours (SigLIP + CUT3R) & 5.91 & 55.0 & 46.7 & 41.8 \\
Ours (+ WM-2D only) & 5.89 & 56.0 & 47.0 & 42.0 \\
Ours (+ WM-3D only) & 5.90 & 55.4 & 47.2 & 41.9 \\
\textbf{Ours (+ WM-2D + WM-3D)} & \textbf{5.82} & \textbf{57.6} & \textbf{48.7} & \textbf{42.9} \\
\hline
\end{tabular}
\end{table}

\begin{table}[!t]
\centering
\caption{Spatial encoder ablation on R2R \texttt{val-unseen}.}
\label{tab:encoder_ablation}
\begin{tabular}{l|c|c c c c}
\hline
Encoder & Time (s) & SR $\uparrow$ & SPL $\uparrow$ & OSR $\uparrow$ & NE $\downarrow$ \\
\hline
VGGT \cite{wang2025vggt} & OOM & OOM & OOM & OOM & OOM \\
InfiniteVGGT \cite{yuan2026infinitevggt} & 0.284 & 43.2 & 38.0 & 54.4 & 6.61 \\
\textbf{Ours (CUT3R)} \cite{wang2025continuous} & \textbf{0.245} & \textbf{48.7} & \textbf{42.9} & \textbf{57.6} & \textbf{5.82} \\
\hline
\end{tabular}
\end{table}

\begin{table}[!t]
\centering
\caption{Performance by task horizon on R2R \texttt{val-unseen}.}
\label{tab:complexity}
\setlength{\tabcolsep}{3pt}
\begin{tabular}{p{1.5cm}|l|r|r r r r}
\hline
Horizon & Model & \#Ep & SR $\uparrow$ & SPL $\uparrow$ & OSR $\uparrow$ & NE $\downarrow$ \\
\hline
\multirow{3}{*}{\parbox{1.5cm}{\centering Short\\(1--50)}} & Baseline & 459 & 51.20 & 48.18 & 55.34 & 5.08 \\
 & Ours & 486 & 51.23 & 48.84 & 54.53 & 4.86 \\
 & \textbf{Difference} & +27 & \textbf{+0.03} & \textbf{+0.66} & -0.81 & \textbf{-0.22} \\
\hline
\multirow{3}{*}{\parbox{1.5cm}{\centering Medium\\(50--100)}} & Baseline & 1038 & 49.61 & 43.79 & 61.27 & 5.64 \\
 & Ours & 1061 & 54.29 & 48.04 & 63.71 & 5.46 \\
 & \textbf{Difference} & +23 & \textbf{+4.68} & \textbf{+4.25} & \textbf{+2.44} & \textbf{-0.18} \\
\hline
\multirow{3}{*}{\parbox{1.5cm}{\centering Long\\($\ge$100)}} & Baseline & 342 & 20.18 & 10.61 & 34.21 & 9.11 \\
 & Ours & 292 & 24.32 & 14.25 & 40.75 & 8.74 \\
 & \textbf{Difference} & -50 & \textbf{+4.14} & \textbf{+3.64} & \textbf{+6.54} & \textbf{-0.37} \\
\hline
\multirow{3}{*}{\parbox{1.5cm}{\centering Overall}} & Baseline & 1839 & 44.54 & 38.72 & 54.76 & 6.15 \\
 & Ours & 1839 & 48.72 & 42.88 & 57.64 & 5.82 \\
 & \textbf{Difference} & 0 & \textbf{+4.18} & \textbf{+4.16} & \textbf{+2.88} & \textbf{-0.33} \\
\hline
\end{tabular}
\par\vspace*{0.25em}
\footnotesize
\textbf{Note.} ``\#Ep'' denotes the number of episodes in each horizon for each model.
\end{table}

\begin{table}[!t]
\centering
\caption{Ablation on attention mask design on R2R \texttt{val-unseen}.}
\label{tab:mask_ablation}
\begin{tabular}{l|c c c c}
\hline
Mask Design & NE$\downarrow$ & OSR$\uparrow$ & SR$\uparrow$ & SPL$\uparrow$ \\
\hline
Leaky & 6.81 & 51.3 & 40.2 & 35.7 \\
w/o Isolation & 6.98 & 51.1 & 39.9 & 35.3 \\
\textbf{Ours} & \textbf{5.82} & \textbf{57.6} & \textbf{48.7} & \textbf{42.9} \\
\hline
\end{tabular}
\end{table}


\begin{figure*}[!t]
\centering
\includegraphics[width=\textwidth]{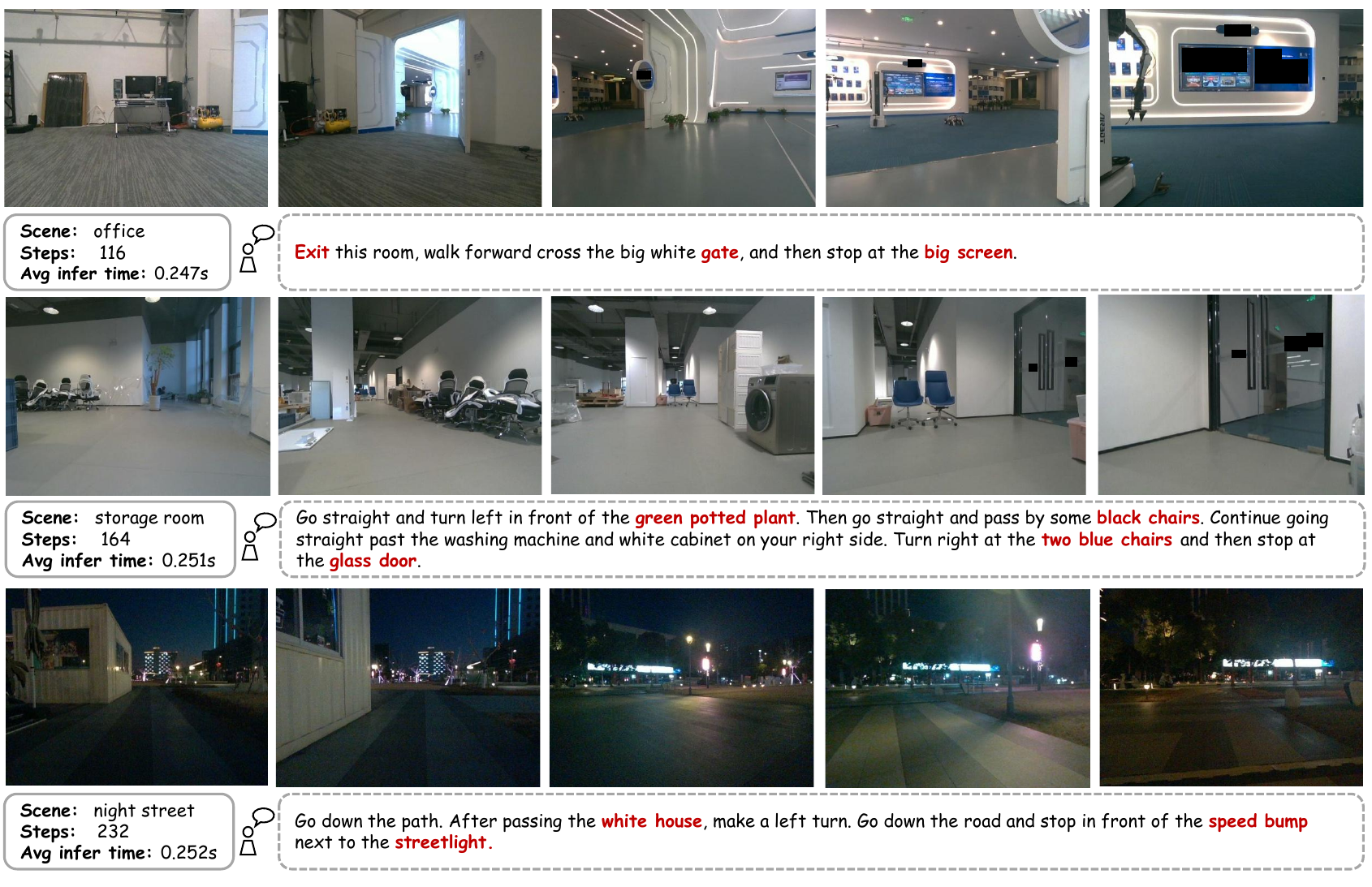}
\caption{First-person views from ARX-Lift2 under diverse indoor/outdoor lighting.}
\label{fig:robot}
\end{figure*}

\begin{table}[!t]
\centering
\caption{Real-robot success rates (completed/total) by scene and lighting.}
\label{tab:robot_success}
\begin{tabular}{l|l|c|c|c}
\hline
Scene & Lighting & NaVid \cite{CIT:Navid} & StreamVLN \cite{wei2025streamvln} & \textbf{Ours} \\
\hline
\textit{Indoor} & & & & \\
\quad Office & Bright & 7/30 & 12/30 & \textbf{20/30} \\
\quad Warehouse & Bright & 6/30 & 12/30 & \textbf{18/30} \\
\quad Corridor & Moderate & 11/30 & 16/30 & \textbf{22/30} \\
\hline
\textit{Outdoor} & & & & \\
\quad Afternoon & Bright & 6/30 & 10/30 & \textbf{18/30} \\
\quad Dusk & Moderate & 4/30 & 6/30 & \textbf{11/30} \\
\quad Night Street & Low & 2/30 & 6/30 & \textbf{9/30} \\
\hline
\end{tabular}
\end{table}

\subsection{Main Results on VLN-CE Benchmarks}
Table~\ref{tab:main} compares PROSPECT with prior methods on VLN-CE R2R/RxR \texttt{val-unseen} under controlled training-data regimes for fair comparison. PROSPECT uses single-view RGB per step (no depth, odometry, or panoramic inputs) and achieves first-tier performance both with MP3D-only navigation data and with scaled training that includes non-MP3D data (e.g., ScaleVLN). Our performance gains on RxR \texttt{val-unseen} are substantially larger than those on R2R (Table~\ref{tab:main}). RxR contains twice as many evaluation episodes as R2R, with a longer average trajectory length (15.32~m vs.\ 9.89~m, i.e., $1.55\times$), and markedly longer instructions (about 120 words on average versus 32, nearly $4\times$). RxR is widely regarded as a more long-horizon and challenging benchmark; the larger improvements on RxR show that the proposed paradigm is particularly beneficial for long-horizon instruction-following navigation.

\subsection{Ablations}

\subsubsection{Module Ablation}
We ablate PROSPECT on R2R/RxR/R2R-EnvDrop under one-epoch SFT, with SigLIP-only as the baseline. Table~\ref{tab:ablation_modules} shows that SigLIP--CUT3R fusion consistently improves performance, and adding either the 2D or 3D latent prediction objective yields additional gains in SR/SPL. Combining both objectives achieves the best result (SR 48.7, SPL 42.9), indicating complementary semantic and geometric predictive signals that jointly provide stronger inductive bias for navigation.

\subsubsection{Spatial Encoder Choice: CUT3R vs. (Infinite)VGGT}
We compare CUT3R with VGGT-style spatial encoders under one-epoch SFT on R2R/RxR/EnvDrop and evaluation on R2R \texttt{val-unseen}. VGGT often OOMs on long R2R episodes (most exceed 30 frames), so we use InfiniteVGGT as the strongest streaming VGGT baseline. Table~\ref{tab:encoder_ablation} shows CUT3R achieves better accuracy and lower latency, which we attribute to its absolute-scale spatial representation versus first-frame-relative scale in VGGT-style encoders.

\subsubsection{Task Complexity: Short vs. Medium vs. Long Horizon}
We stratify R2R \texttt{val-unseen} by executed steps: short (1--50), medium (50--100), long ($\ge 100$). Table~\ref{tab:complexity} shows that PROSPECT matches baseline on short tasks and yields larger improvements on medium and long tasks, indicating stronger generalization under long streaming context.

\subsubsection{Ablation on Query Attention Mask Design}
To validate our query attention mask design, we conduct ablations on three variants evaluated on R2R \texttt{val-unseen}. \textbf{Ours} enforces strict causal masking and full 2D/3D query isolation, preventing each query from attending to future navigation tokens or queries of a different modality. \textbf{w/o Isolation} retains the causal constraint but allows 2D and 3D queries within the same round to attend each other, introducing cross-modal feature entanglement. \textbf{Leaky} applies a standard causal mask without isolation, enabling queries to implicitly access future navigation tokens and causing information leakage during training. Results in Table~\ref{tab:mask_ablation} confirm that both isolation and causal strictness are essential for robust navigation performance.

\subsection{Real-Robot Results}
We evaluate on ARX-Lift2 across indoor/outdoor scenes with varying illumination. Per scene, we group episodes into short/medium/long horizons by the number of executed steps ($<50$, $[50,100)$, and $\ge 100$). For each horizon, we design five distinct instructions and execute each instruction twice, resulting in 30 trials per scene. All real-world scenes are unseen during training; we select goal locations and trajectories to cover diverse layouts and visual appearances (e.g., texture, clutter, illumination) and varying instruction complexity. Success requires reaching within 0.3\,m of the goal within 500 steps and outputting \texttt{STOP}; collisions count as failures. The indoor trials are under ceiling lighting and are well-lit; the outdoor trials span afternoon, dusk, and night.

Table~\ref{tab:robot_success} shows PROSPECT improves over NaVid and StreamVLN across scenes and lighting. Fig.~\ref{fig:robot} visualizes example runs; additional results appear in the supplementary video.

\section{Conclusion}
We presented PROSPECT, a unified streaming VLN agent integrating streaming VLA, CUT3R-based absolute-scale 3D encoding, and latent predictive representation learning via stream query tokens. The predictive branch is used only during training to shape representations without inference overhead. PROSPECT achieves first-tier VLN-CE performance and robust real-robot navigation under diverse lighting. Future work will explore further refinements to robustness and efficiency in real-world deployments.

\FloatBarrier
\bibliographystyle{IEEEtran}
\bibliography{main}

\balance
\end{document}